\documentclass{ifacconf}

\usepackage{graphicx}
\usepackage{natbib}
\usepackage{multicol}
\usepackage{amssymb}
\usepackage{amsmath}

\begin{document}
\begin{frontmatter}

\title{An Evaluation of Classification Methods for 3D Printing Time-Series Data\thanksref{footnoteinfo}} 
% Title, preferably not more than 10 words.

\thanks[footnoteinfo]{This publication has resulted from research supported in part by a  grant from Science Foundation Ireland (SFI) under Grant Number 16/RC/3872 and is co-funded under the European Regional Development Fund.}

\author[First]{Vivek Mahato} 
\author[Second]{Muhannad Ahmed Obeidi} 
\author[Second]{Dermot Brabazon}
\author[First]{P\'{a}draig Cunningham}

\address[First]{I-Form SFI Research Centre for Advanced Manufacturing, University College Dublin, Ireland
(e-mail: author@ i-form.ie).}
\address[Second]{I-Form SFI Research Centre for Advanced Manufacturing, Dublin City University, Ireland
(e-mail: author@ i-form.ie).}

\begin{abstract}                % Abstract of not more than 250 words.
Additive Manufacturing presents a great application area for Machine Learning because of the vast volume of data generated and the potential to mine this data to control outcomes. In this paper we present preliminary work on classifying infrared time-series data representing melt-pool temperature in a metal 3D printing process. Our ultimate objective is to use this data to predict process outcomes (e.g. hardness, porosity, surface roughness). 
In the work presented here we simply show that there is a signal in this data that can be used for the classification  of different components and stages of the AM process.
In line with other Machine Learning research on time-series classification we use $k$-Nearest Neighbour classifiers. 
The results we present suggests that Dynamic Time Warping is an effective distance measure compared with alternatives for 3D printing data of this type. 
\end{abstract}


\end{frontmatter}
%===============================================================================

\section{Introduction}

\begin{figure}
\begin{center}
\includegraphics[width=8cm]{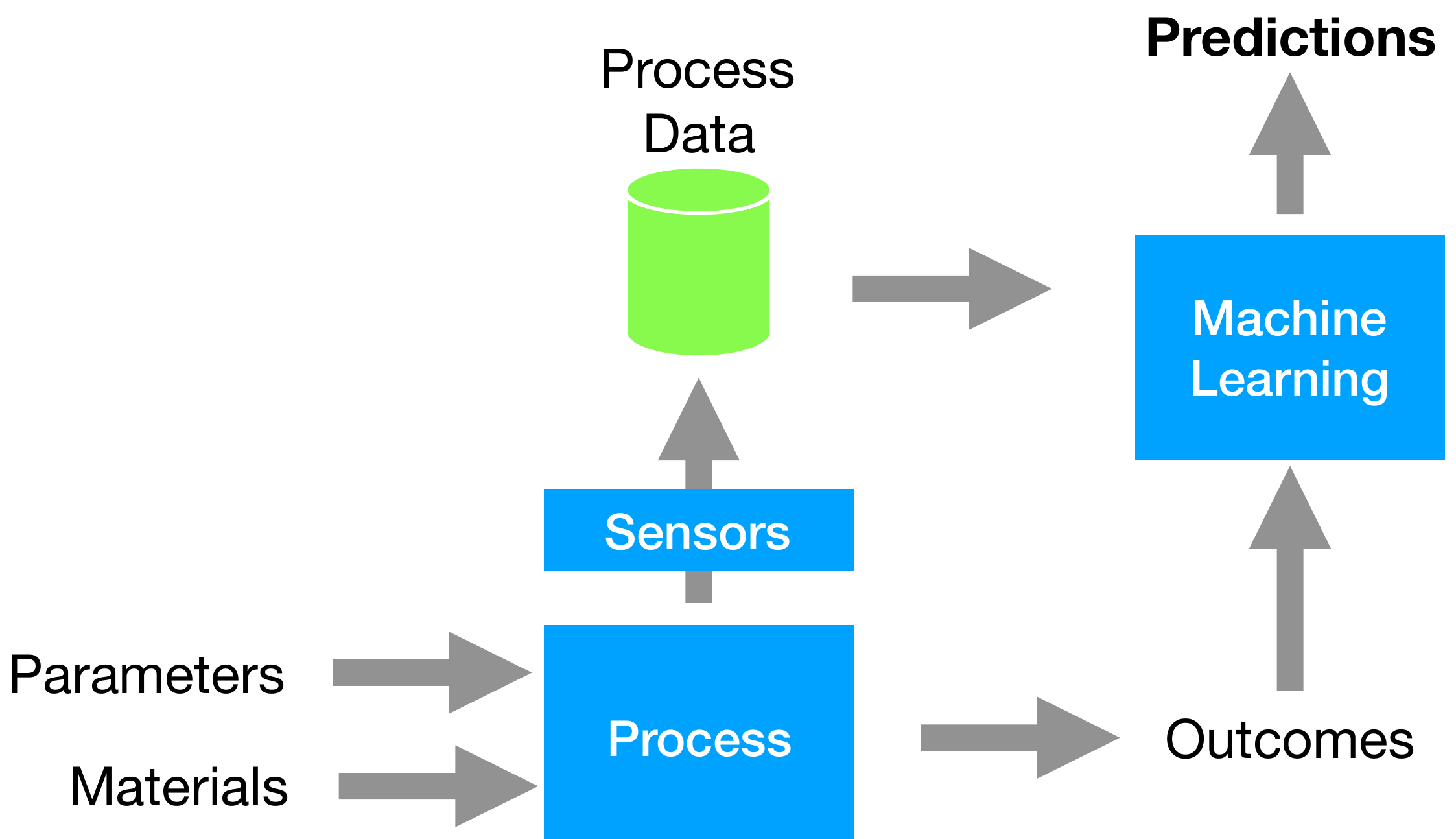}    % The printed column width is 8.4 cm.
\caption{Supervised Machine Learning in AM.} 
\label{fig:process}
\end{center}
\end{figure}

It is generally accepted that one of the big challenges with Additive Manufacturing (AM) is that the processing parameters are difficult to tune
\citep{qi2019applying}. On the positive side, AM produces vast amounts of data that can be mined to address this problem. A strategy for using Supervised Machine Learning in AM is shown in  Fig.~\ref{fig:process}. The objective is to analyse data gathered through sensors during the process and analyse that data to predict process outcomes. In the context of AM, these outcomes might relate to product quality, for instance hardness, surface roughness or porosity. 

It is in the nature of AM that the data coming from the sensors is likely to be in the form of a time-series. The time-series data might be aggregated into feature vector representations so that standard Machine Learning (ML) methods can be applied; however one of our objectives is to work with the raw time-series data coming from the process. 

Our overall objective is to use pyrometers to measure meltpool temperature and mine this data to predict process outcomes. The hypothesis is that anomalies such as pores will have characteristic signatures in the temperature time-series. Our first step in this direction is to find out how much signal there is in the pyrometer data. 

We have data from the printing of 5mm sample cubes printed in batches of 27. We consider two different ML tasks on this data:
\begin{itemize}
    \item Can we distinguish different cubes based on the temperature time-series?
    \item Can we distinguish different layers in the print process? 
\end{itemize}

The details about the data and the specifics of these classification tasks are presented in the next section. The time-series classification methods we evaluate are described in section \ref{sec:sim}. The results of the evaluation are presented in section \ref{sec:eval} and our next steps are outlined in section \ref{sec:future}. 

\section{Context}\label{sec:context}
The analysis presented here uses data from the Aconity MINI 3D printer.\footnote{https://aconity3d.com/products/aconitymini/}
This is a powder-bed fusion (PBF) printer using selective laser melting (SLM). While AM using PBF has great potential, the process is very sensitive to input parameters and there is a lot that can go wrong during a build. Common problems are voids, pores and lack of fusion due to spatter or under or over melting in the meltpool \citep{ozel2018process}.  

PBF operates in an inert gas environment and the flow of this gas can have an impact on cooling, particularly for high gas flow rates or when a turbulent flow occurs. Excessive amounts of powder may be used within the process if it is not well controlled. There is a strong incentive to recycle powder but potential problems due to recycled powder quality can prevent this \citep{gorji2020new}. Degradation of the powder feedstock due to repeated recycling may negatively impact on the mechanical properties of components built using this powder.

Our objective is to be able to identify these problems as they arise during the build process. In this paper we assess the potential to do this by analysing pyrometer data that tracks the temperature of the melt pool. 

\subsection{3D Printing Data}

%The data is gathered from test builds on an Aconity MINI 3D printer.\footnote{https://aconity3d.com/products/aconitymini/}
The temperature data is recorded by means of two pyrometers from KLEIBER Infrared GmbH. The two pyrometers detect the heat emission light in the range of 1500 to 1700nm via two detectors. The measured light which is reflected from the meltpool area, is split into two paths by means of optical filters and transmitted to the pyrometers via optical fibre cables. In the Aconity 3D printer used in the work, the scanner and the pyrometers are configured to cover $x$ and $y$ values (for each layer) in the range of 0 to 32768 bit covering an area of 400$\times$400 mm. This results in a calibration value of 81.92 bit/mm. Taking in account the pyrometer frequency of 100 kHz, this will produce a response of one single measurement of the meltpool temperature in every 10$\mu$s; e.g. one measurement in each ten microns in the $x$ and $y$ directions based on a scanning speed of 1000 mm/s. 

\begin{figure}
\begin{center}
\includegraphics[width=8cm]{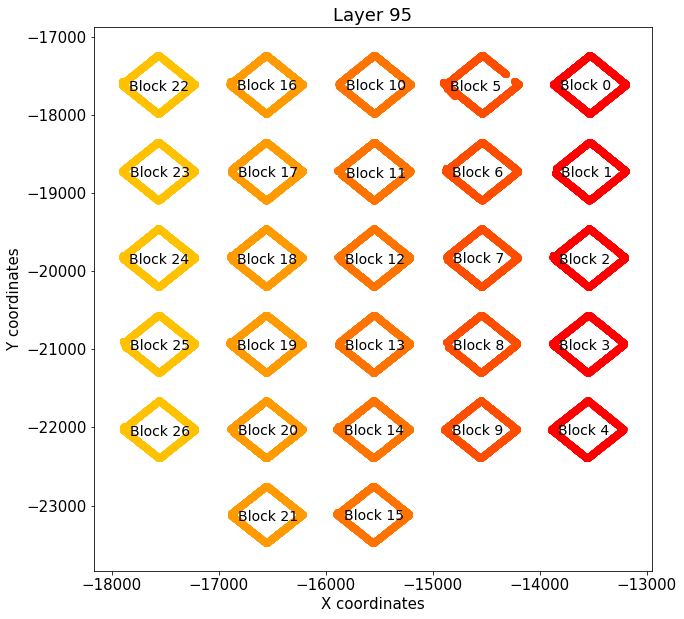}    
\caption{Perimeter scans from a sample layer of the 27 block build.} 
\label{fig:scan}
\end{center}
\end{figure}

\begin{figure}
\begin{center}
\includegraphics[width=8cm]{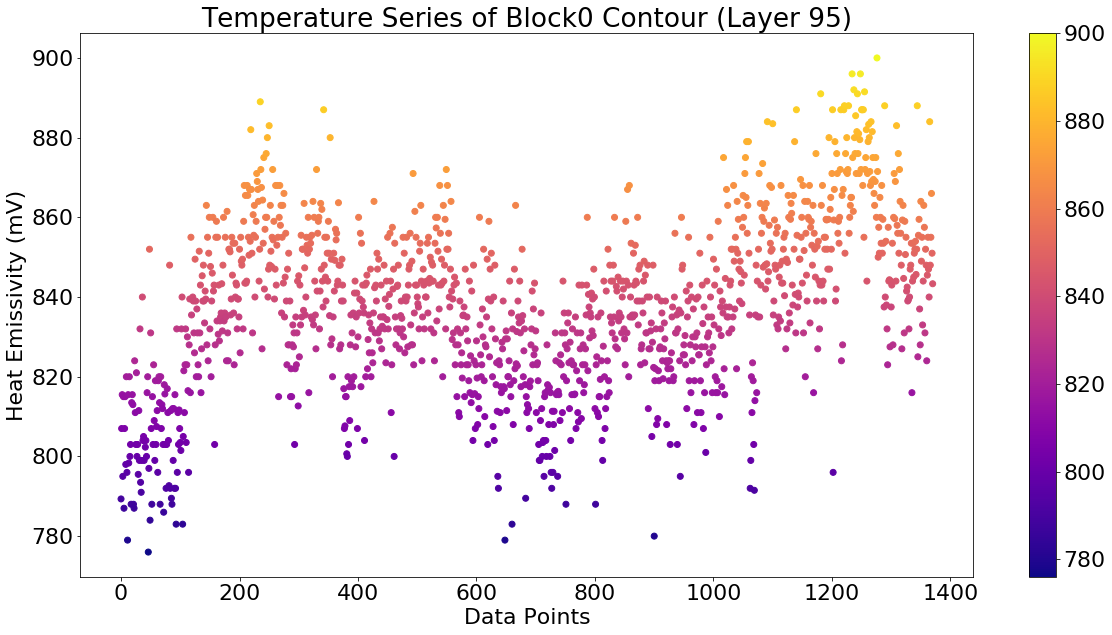}    
\caption{The temperature time-series of a typical Block 0 perimeter.} 
\label{fig:tempseries}
\end{center}
\end{figure}

The data analysed here comes from a build of 27 $\times$ 5mm blocks built with a layer thickness of 20$\mu$m resulting in 250 layers in the build (see Fig. \ref{fig:scan}). The temperature data for each layer (all 27 blocks) contains approximately 700,000 data points. The laser scan for each block comprises the perimeter scan shown in  Fig. \ref{fig:scan} and a raster scan to fill in the interior of the block. A sample temperature time-series for a perimeter scan for Block 0 is shown in Fig. \ref{fig:tempseries}. These values are the emissivity measured in (mV) which are directly  proportional to the melting temperature. 

\begin{figure}
\begin{center}
\includegraphics[width=8cm]{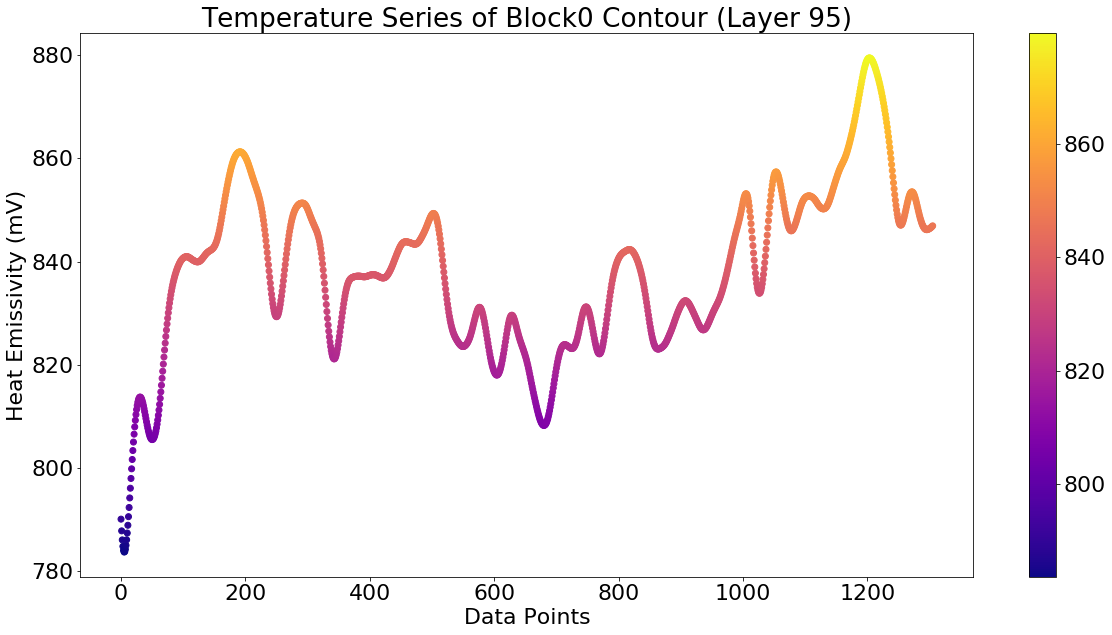}    
\caption{The time-series in Fig. \ref{fig:tempseries} after passing through a Butterworth filter.}
\label{fig:filttempseries}
\end{center}
\end{figure}

Because these temperature readings are evidently noisy, we also consider a smoothed version of the time-series. The same time-series passed through a Butterworth filter is shown in Fig. \ref{fig:filttempseries}.\footnote{Data will be available for download through the Time Series Classification data repository https://timeseriesclassification.com.}

\subsection{The Classification Tasks}\label{sec:class_tasks}
The Aconity printer uses Argon gas to maintain an inert environment. The gas flows right to left across the build plate as shown in Fig. \ref{fig:scan}. For the first classification tasks we consider the perimeter scan of a block as a temperature time-series. We consider:
\begin{itemize}
    \item Block 0 versus Block 22 (easy),
    \item Block 0 versus Block 1 (hard),
    \item Block 1 versus Block 22 (easy).
\end{itemize}
Each block is represented by 250 samples, one for each layer. The parameter tuning is done by 6-fold cross-validation over the bottom 212 layers, and then the model is evaluated over the top 38 layers that have been held back from model selection and training.

The second classification task is to distinguish between the top 10 and bottom 10 layers. With 27 blocks there are 270 samples in each class. The evaluation is done by holding back data from 8 blocks for testing.  

% \begin{figure*}
% \begin{multicols}{3}
%     \includegraphics[width=\linewidth]{images/dtw-blank.png}
%      \par 
%     \includegraphics[width=\linewidth]{images/dtw-linear.png}\par 
%     \includegraphics[width=\linewidth]{images/dtw-warp.png}\par
% \end{multicols}
% \caption{Dynamic Time Warping; (a) Two similar time-series displaced in time. (b) Mapping of data-points without warping. (c) An example of DTW non-linearly mapping two time-series displaced in time \citep{mahato2019scoring}. }
% \label{fig:dtw}%
% \end{figure*}

 \begin{figure*}
     \centering
     \includegraphics[width=0.9\linewidth]{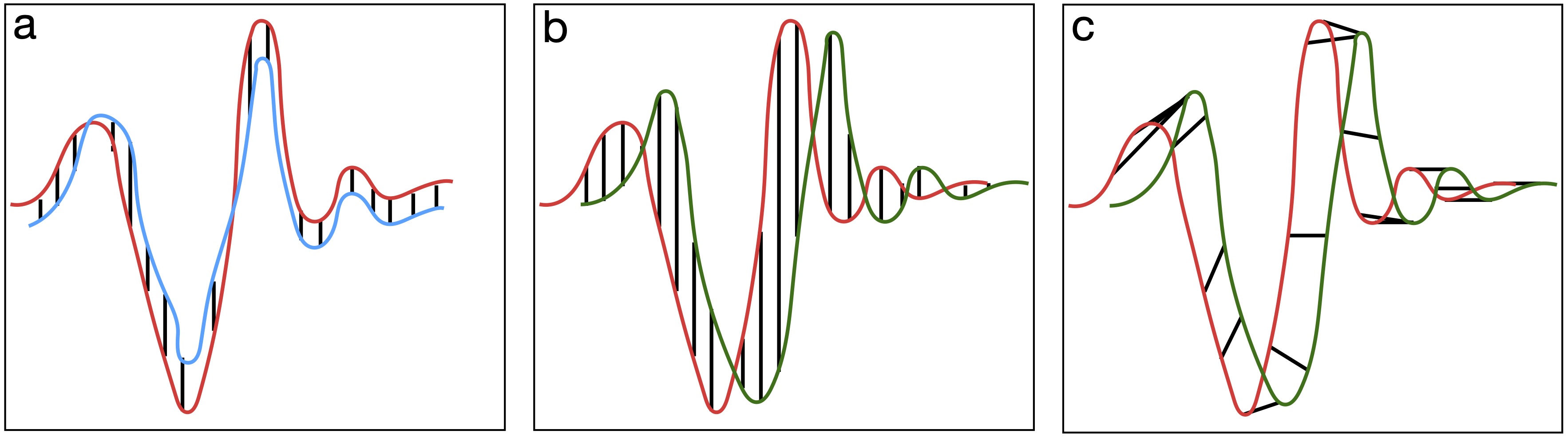}
 \caption{Dynamic Time Warping; (a) Two similar time-series. (b) Two similar time-series displaced in time -- Euclidean distance is large. (c) An example DTW  mapping for the two time-series in (b).}
 \label{fig:dtw}%
 \end{figure*}

\section{Machine Learning for Time-Series}\label{sec:sim}

We are dealing with time-series data of specific characteristics. When it comes to time-series classification $k$-Nearest Neighbor ($k$-NN) has a special status, which may be because some popular ML methods such as Decision Trees and Support Vector Machines will not work with time-series data. There is a presence of several similarity/distance methods that developed to applied with $k$-NN to temporal data, which makes it more efficient. We examine three such methods in our evaluation: 
\begin{itemize}
    \item Dynamic Time Warping (DTW),
    \item Symbolic Aggregate approXimation (SAX),
    \item Symbolic Fourier Approximation (SFA).
\end{itemize}
These methods are described in the subsections that follow, and then we present the results.

\subsection{Dynamic Time Warping}\label{sec:dtw}
Euclidean distance is a popular metric for assessing similarity between feature value representations.
Euclidean distance will work well with time-series data if the time-series are well aligned (see Figure \ref{fig:dtw}(a)). However, a small misalignment will result in a large Euclidean distance (Figure \ref{fig:dtw}(b)).
Nevertheless, Euclidean distance is included as a baseline in our experiments. As the name suggests, DTW attempts to address this misalignment by allowing more flexible mapping in the time dimension ((Figure \ref{fig:dtw}(c)). The DTW distance is defined as follows:

\begin{equation}
    DTW(\mathbf{q,x})=\underset{\pi}{\text{min}}\sqrt{\sum_{(i,j)\in\pi}{d(q_i,x_j)^2}}
\end{equation}

where $\pi=[\pi_1,...,\pi_l,...,\pi_L]$ is the optimum path (mapping) having the following properties:
\begin{itemize}
    \item $m = |\mathbf{q}|,n = |\mathbf{x}|$
    \item $\pi_1 =(1,1), \pi_L = (m,n)$
    \item $\pi_{l+1} - \pi_l \in\{(1,0),(0,1),(1,1)\}$
\end{itemize}

A cost matrix is constructed by DTW, where each cell $(i,j)$ contains the distance between $q_i$ and $x_j$. The overall distance is the sum of distances taken by the shortest path through the grid. The extent of the deviation from the main diagonal reflects the \emph{warping}. The computational complexity of DTW is $O(n,m)$ because it entails a search through the matrix. This complexity is effectively $O(n^2)$ in the length of the time-series -- so DTW is computationally expensive. To improve the performance of DTW, and reduce its time and memory complexity, we utilise the Sakoe-Chiba \citeyearpar{sakoe_chiba} global constraint in the model.

\begin{figure*}
% \begin{multicols}{3}
%     \includegraphics[width=\linewidth]{images/raw.jpg}
%      \par 
%     \includegraphics[width=\linewidth]{images/paa.jpg}\par 
%     \includegraphics[width=\linewidth]{images/sax.jpg}\par
% \end{multicols}

     \centering
     \includegraphics[width=0.9\linewidth]{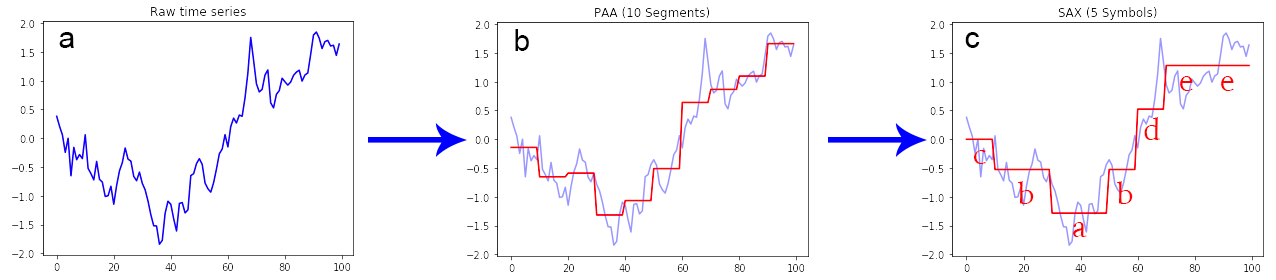}
\caption{Symbolic Aggregate Approximation; The raw time-series in (a) will be represented by the sequence \textsf{cbabdee} in (c) \cite{DBLP:conf/aics/MahatoOC18}.}%
    \label{fig:sax}%
\end{figure*}

\subsection{Symbolic Aggregate approXimation}\label{sec:sax}
In the past few decades, there was much research around developing symbolic representations of time-series data. The idea is to harness the power of text processing algorithms to solve time-series tasks. A summary of such methods is provided by \cite{lin_keogh_lonardi_chiu_2003}.

Symbolic Aggregate Approximation (SAX) is one such algorithm that coverts a time-series into a series of symbols and obtains dimensionality and numerosity reduction (i.e. more compact representation) of the original time-series. These transformations present a distance measure that is lower bounding on corresponding measures on the original series \citep{lin_keogh_lonardi_chiu_2003}.

\subsubsection{Piecewise Aggregate Approximation}
SAX employs Piecewise Aggregate Approximation (PAA) for dimensionality reduction of the original time-series. PAA achieves this by slicing the time-series into bins of equal sizes. The series can then be represented by the mean values in these bins (the PAA coefficients).

Consider a time series $S$ of length $n$.  PAA reduces the series $S$ from size $n$ to $m$, where $m \leq n$. It achieves this reduction by transforming $S$ into a vector $\bar{S}=(\bar{s}_{1},\bar{s}_{2},... ,\bar{s}_m )$, where each of $s_i$ is computed as follows:
\begin{equation}
    \bar{s}_i = \frac{m}{n} \sum _{j=\frac{n}{m}(i-1)+1}^{\frac{n}{m}i} s_j
\end{equation}

With PAA there are two exceptional situations worth noting \citep{keogh_pazzani_2000}:
\begin{itemize}
    \item $m=n$, the transformation is similar to the original time-series.
    \item $m=1$, the transformation is the mean of the original time-series.
\end{itemize}
Due to the difficulty in comparing two time-series of different scales, SAX normalizes the original series so that the mean is zero and standard deviation is one, before passing to PAA for transformation \citep{lin_keogh_lonardi_chiu_2003,keogh_kasetty_2002}.

SAX passes the PAA transformed series through another discretisation procedure that converts them into symbols. SAX achieves this conversion by discretising the levels into $a$ bins of approximately equal size. These discretised levels typically follow a Gaussian distribution, so these bins get bigger further from the mean. 
The discretised bins are separated by breakpoints that forms a sorted list $B = \beta_1, ..., \beta_a{}_{-1}$, in a way that the area under a $N(0,1)$ Gaussian curve from $\beta_i$ to $\beta_{i+1}$ = $\frac{1}{a}$, where $\beta_0$ and $\beta_a$ are $-\infty$ and $\infty$ respectively \citep{lin_keogh_lonardi_chiu_2003}.

SAX is provided with a pool of symbols $S = (s_1,s_2, ..., s_m)$, where $m$ is the size of the pool. After computing all the breakpoints, SAX performs the symbolic transformation as follows. First, the algorithm transforms the original series into a series of PAA coefficients. We take the smallest breakpoint $\beta_1$ first and map all the PAA coefficients that are less than $\beta_1$ to the symbol $s_1$. Then we take $\beta_2$ (the second smallest) breakpoint, and all the PAA coefficients that fall between $\beta_1$ and $\beta_2$ gets mapped to $s_2$, and so on until all the PAA coefficients get mapped to its corresponding symbol.

The SAX algorithm includes a module that uses a sliding window technique with an adjustable size. The idea is to extract all the symbols inside the window and concatenate them to create a SAX word. The sliding window then shifts to its right, and extract the symbols in this new window to create another SAX word, and this goes on until the sliding window hits the last frame to create the last word. This collection of words also known as a ``bag-of-words", represents the original time-series.

After the transformation of all the time-series data in our dataset, we can easily calculate the distance between any two given time-series by using any string metric on their symbolic representation. Levenshtein distance  \citep{yujian2007normalized} is one such example of a string metric that is popular.

\subsection{Symbolic Fourier Approximation}\label{sec:sfa}

SFA \citep{sc_2012} is another example of an algorithm built on the idea of dimensionality reduction by symbolic transformation. SAX tries to keep the data in the time-domain, whereas SFA transforms the data to bring it in the frequency domain because, in the frequency domain, each dimension has approximate information of the entire series. One can also enhance the overall quality of the estimate by increasing the dimensions. Working with the time-domain requires deciding the length of approximation in advance, and a prefix of this length factors just a subset of the time-series \citep{sc_2012}.

\begin{figure}[ht]
    \centering
    \includegraphics[width=8cm]{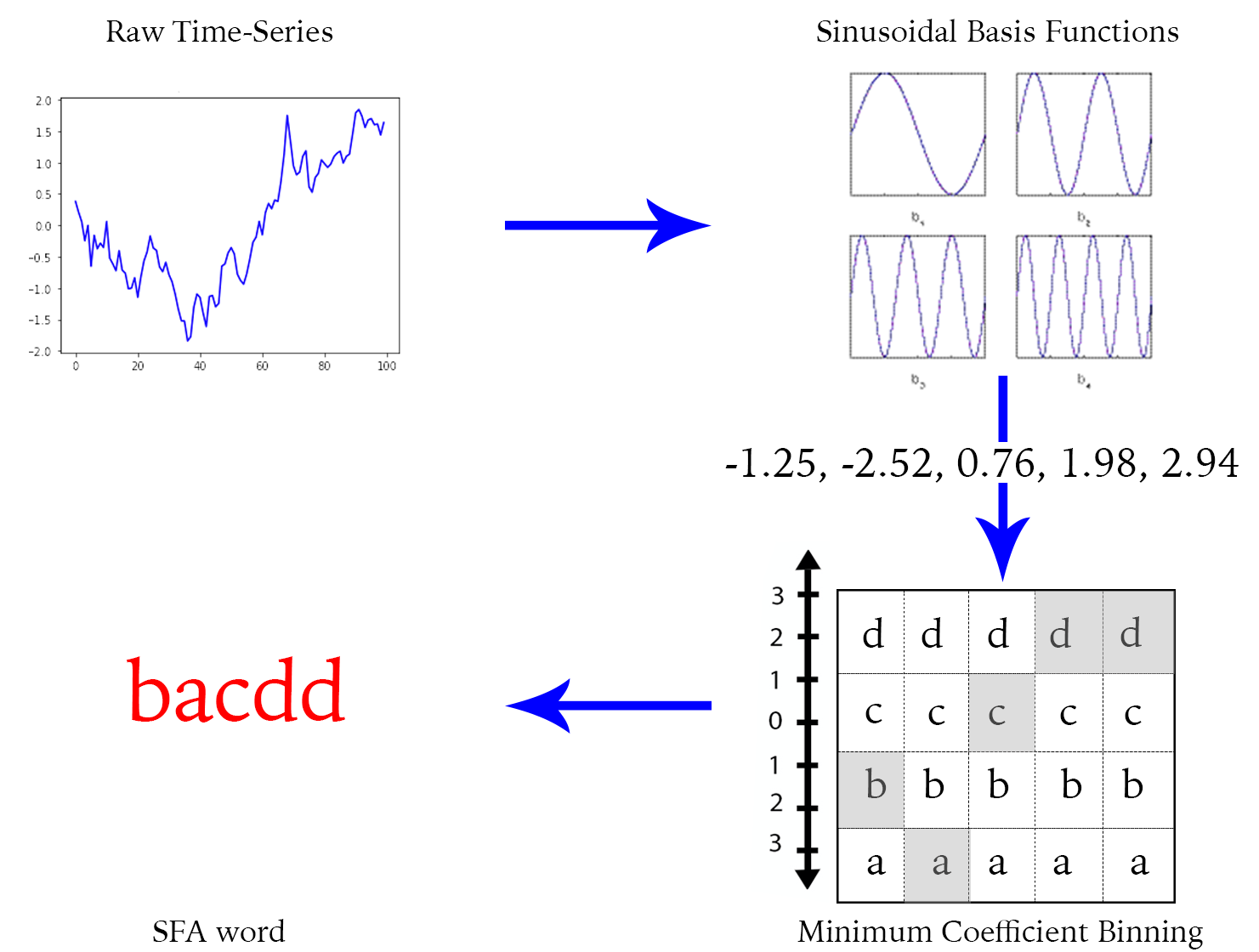}
    \caption{Symbolic Fourier Approximation; The raw time-series will be represented by the sequence \textsf{bacdd} \citep{DBLP:conf/aics/MahatoOC18}.}
    \label{fig:sfa}
\end{figure}

\subsubsection{Discrete Fourier Transform}
SFA employs Discrete Fourier Transformation (DFT) as its dimensionality reduction technique to focus the data in the frequency domain, rather than PAA used in SAX. DFT and the continuous Fourier Transform for signals are alike, and are known only at $N$ instants by sample times $T$, which is a finite series of data.

Let $S(t)$ be the continuous signal which is the source of the data. Let N samples be denoted $s[0], s[1], ..., s[N-1]$.
The Fourier transformation of the original signal, $S(t)$, would be:
\begin{equation}
    F(\omega_{k}) \triangleq \sum_{n=0}^{N-1}s(t_n)e^{-j \omega_k t_n} ,     k=0,1,2,...,N-1
\end{equation}

To determine the signal's frequency content at $S[k]$, DFT examines a time-domain signal at $s(n)$ by comparing it against sinusoidal basis functions through correlation. The first few basis functions describe the gradual changing regions, while the later basis functions describe the rapid changes like gaps and noise. Therefore, using only the first few basis functions, one can have a good approximation of the entire series \citep{sc_2012}.

The SFA model uses DFT approximation as part of preprocessing by transforming the original time-series into a series of DFT coefficients. SFA then applies Multiple Coefficient Binning (MCB) method for computing multiple discretisations of the coefficients series (see Figure \ref{fig:sfa}). 
MCB engages in the mapping of the DFT coefficients to their respective symbols and then concatenates them to form an SFA word, transforming the time-series into a symbolic representation.

SFA also uses a sliding window in the same manner as SAX and the output is also a ``bag-of-words" representation of the time-series.

\section{Evaluation}\label{sec:eval}
In the evaluation we consider the two classification tasks described in section \ref{sec:class_tasks}. For convenience we refer to these as `Up Wind versus Down Wind' and `High versus Low'. For the tasks we consider two versions of the data, the raw data and the filtered data as shown in Fig. \ref{fig:filttempseries}.
In addition to the three specialised time-series methods presented in section \ref{sec:sim} we also consider Euclidean distance and the Mean of the time-series as baselines. The Mean is included to show that very simple aggregate statistics are not sufficient. 

\subsection{Up Wind versus Down Wind}

The results on the first set of tasks are shown in Tables \ref{tab:updown} and \ref{tab:fupdown}. It is clear that the Mean and Euclidean baselines do not perform well and DTW appears to be the overall winner. 

\begin{table}[hb]
\caption{Up Wind versus Down Wind (Raw)}
\centering
\begin{tabular}{ccccc}
Model & 0 vs 22 & 0 vs 1 & 1 vs 22 & 0 vs 1 vs 22 \\\hline
Mean & 55.26 & 50.0 & 51.32 & 33.33 \\
Euclidean & 77.63 & 64.47 & 64.47 & 55.26 \\
DTW & 86.84 & 89.47 & 94.74 & 80.70 \\ 
SAX & 80.26 & 61.84 & 65.79 & 57.02 \\ 
SFA & 82.89 & 56.58 & 67.11 & 52.63 \\ \hline
\end{tabular}
\label{tab:updown}
\end{table}

\begin{table}[ht]
\caption{Up Wind versus Down  (Filtered)}
\centering
\label{tab:fupdown}
\begin{tabular}{ccccc}
Model & 0 vs 22 & 0 vs 1 & 1 vs 22 & 0 vs 1 vs 22 \\\hline
Mean & 59.21 & 52.63 & 51.32 & 33.33 \\
Euclidean & 73.68 & 61.84 & 67.11 & 56.14 \\
DTW & 88.16 & 65.79 & 86.84 & 64.04 \\ 
SAX & 82.89 & 57.89 & 84.21 & 59.65 \\ 
SFA & 71.05 & 57.89 & 76.32 & 45.61 \\ \hline
\end{tabular}
\end{table}

\subsection{High versus Low}
The same evaluation is repeated for the High versus Low task. In this case there is only one task so the results are presented in a single table -- Table \ref{tab:highlow}. Again, DTW is the clear winner. It is interesting to note that, whereas Mean was no better than random guessing in the first exercise, it has some classification power here. 

\begin{table}[t]
\begin{center}
\caption{High versus Low}
\label{tab:highlow}
\begin{tabular}{ccc}
Model & Raw & Filtered \\\hline
Mean & 71.25 & 74.38 \\
Euclidean & 67.5 & 68.75 \\
DTW & 89.38 & 90.63 \\ 
SAX & 59.38  & 59.38\\ 
SFA & 43.75  & 54.38 \\ \hline
\end{tabular}
\end{center}
\end{table}

\begin{figure}[ht]
    \centering
    \includegraphics[width=8.5cm]{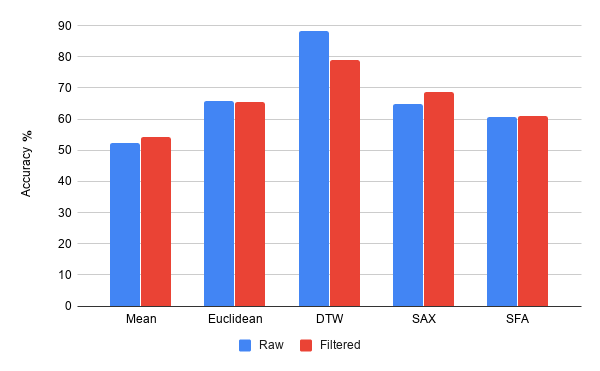}
    \caption{Average model accuracy across all tasks.}
    \label{fig:summary}
\end{figure}

\subsection{Discussion}\label{sec:disc}
Fig. \ref{fig:summary} shows the average model accuracy across all tasks.
There are a few clear conclusions that can be drawn:
\begin{itemize}
    \item It is pretty clear that Euclidean distance is not a competitive distance measure on this data. This is because of the variable length of the time-series and the associated problems of alignment. 
    \item Of the three methods that are specifically conceived to deal with time-series, it seems clear that DTW beats both SAX and SFA. This is probably due to the global nature of DTW that considers the whole time-series whereas SAX and SFA work well when specific signatures in sub-regions of the time-series are important \citep{DBLP:conf/aics/MahatoOC18}.
    \item It is interesting to note that DTW works better with the raw data whereas SAX and SFA perform best with the data that has passed through the Butterworth filter. 
    
\end{itemize}

\section{Conclusion \& Future Work}\label{sec:future}

This evaluation  shows that the temperature time-series data does capture information about the AM process. It also shows that ML methods specialised for time-series analysis are required to get the most from the data. 
As a next step we are carrying out CT scans on the blocks to identify pores and then see if these pores have a characteristic signature in the temperature time-series. 

\begin{ack}
This publication has resulted from research supported in part by a  grant from Science Foundation Ireland (SFI) under Grant Number 16/RC/3872 and is co-funded under the European Regional Development Fund.
\end{ack}

\bibliography{ifac2020}             

\end{document}